\documentclass[a4paper,12pt]{article}

\usepackage[dvipsnames]{xcolor}
\usepackage{amsmath}
\usepackage{amssymb}
\usepackage{dsfont}
\usepackage[english]{babel}
\usepackage[utf8]{inputenc}
\usepackage{times}
\usepackage{graphics}
\usepackage{graphicx}
\usepackage[T1]{fontenc}
\usepackage[official,right]{eurosym}
\usepackage{url}
\usepackage{eso-pic}
\usepackage{tikz}
\usetikzlibrary {arrows.meta,bending,positioning}

\newtheorem{deff}{Definition}[section]
\newtheorem{claim}{Claim}

\newtheorem{prop}{Proposition}[section]
\newtheorem{remark}{Remark}[section]

\newtheorem{example}{Example}[section]
\def\qed{\hbox to\hsize{\hfill\vrule height 1.6ex width 1.5ex depth -.1ex}}

\title{What is a decision problem?}
\author{Alberto Colorni$^\dag$, Alexis Tsouki\`as$^\ddag$ \\ $^\dag$ POLIEDRA, Politecnico di Milano \\ $^\ddag$CNRS-LAMSADE, PSL, Universit\'e  Paris Dauphine}
\date{}

\begin{document}

\thispagestyle{empty}

\enlargethispage*{8cm}
 \vspace*{-38mm}

\AddToShipoutPictureBG*{\includegraphics[width=\paperwidth,height=\paperheight]{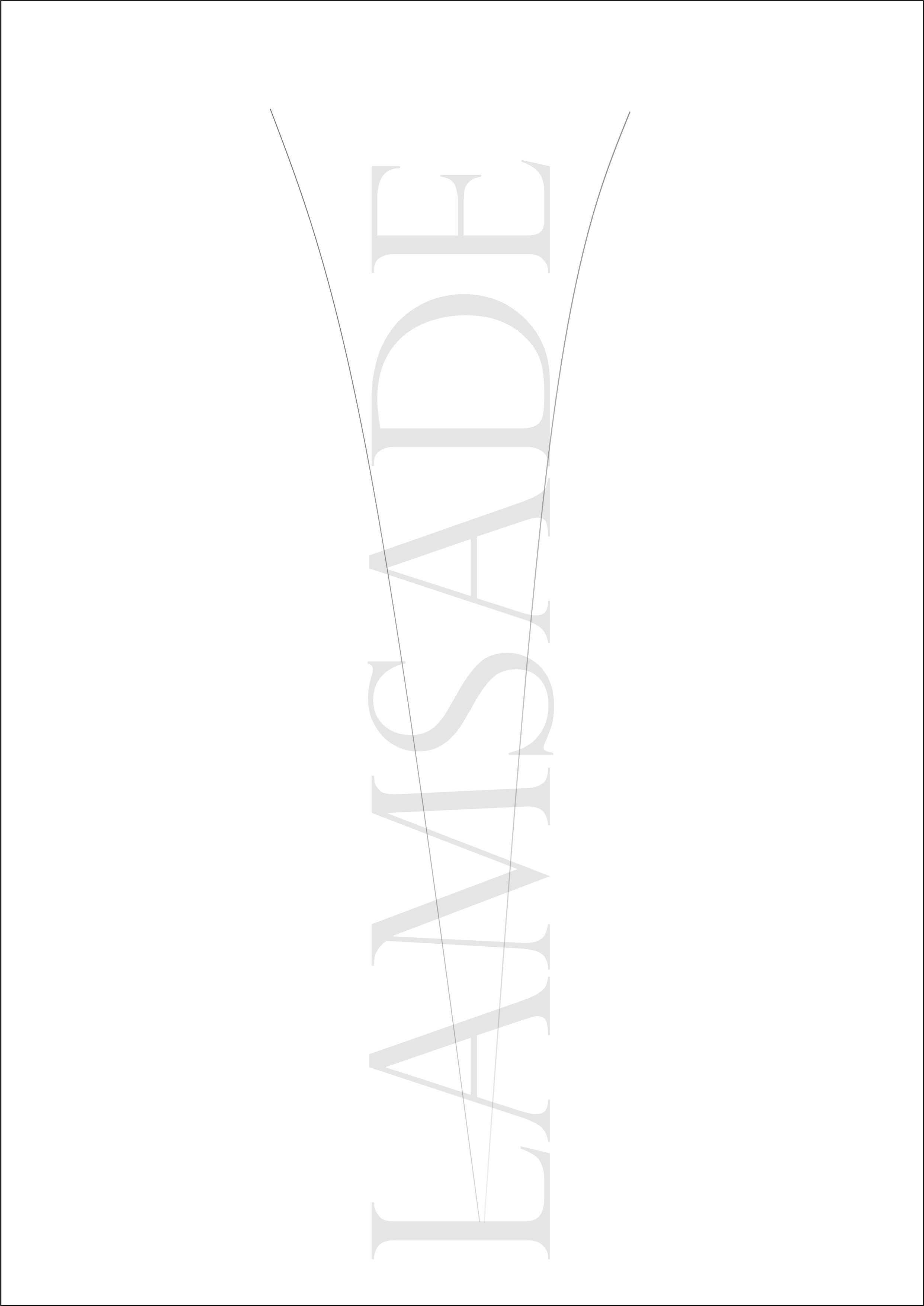}}

\begin{minipage}{24cm}
 \hspace*{-28mm}
\begin{picture}(500,700)\thicklines
 \put(60,670){\makebox(0,0){\scalebox{0.7}{\includegraphics{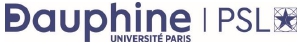}}}}
 \put(60,70){\makebox(0,0){\scalebox{0.3}{\includegraphics{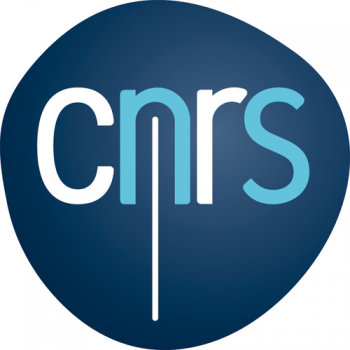}}}}
 \put(320,350){\makebox(0,0){\Huge{CAHIER DU \textcolor{BurntOrange}{LAMSADE}}}}
 \put(140,10){\textcolor{BurntOrange}{\line(0,1){680}}}
 \put(190,330){\line(1,0){263}}
 \put(320,310){\makebox(0,0){\Huge{\emph{404}}}}
 \put(320,290){\makebox(0,0){January 2023}}
 \put(320,210){\makebox(0,0){\Large{What is a decision problem?}}}
 \put(320,100){\makebox(0,0){\Large{Alberto Colorni, Alexis Tsouki\`as}}}
 \put(320,670){\makebox(0,0){\Large{\emph{Laboratoire d'Analyse et Mod\'elisation}}}}
 \put(320,650){\makebox(0,0){\Large{\emph{de Syst\`emes pour l'Aide \`a la D\'ecision}}}}
 \put(320,630){\makebox(0,0){\Large{\emph{UMR 7243}}}}
\end{picture}
\end{minipage}

\newpage

\addtocounter{page}{-1}

\maketitle

\abstract{This paper presents a general framework about what is a decision
problem. Our motivation is related to the fact that decision analysis and operational research are structured (as disciplines) around classes of methods, while instead we should first characterise the decision problems our clients present us. For this purpose we introduce a new framework, independent from any existing method, based upon primitives provided by (or elicited from) the client. We show that the number of archetypal decision problems are finite and so the archetypal decision support methods.}

\newpage

%
%
%
%

%
%
%
\section{Introduction}

The reader should be aware that this paper does not address the title question in a comprehensive way. The problem of what is a decision and what is a decision problem has been addressed in philosophy, psychology and the cognitive sciences, economy, political science etc.. We are not going to make a survey of this, for the rest, extremely interesting literature which is out of the scope of the paper. The reader interested in these aspects can have a look to a number of fundamental texts such as \cite{Habermas81}, \cite{Jeffrey65}, \cite{Kaplan1996}, \cite{Rescher69}, \cite{rosen1book89}, \cite{Simon47}, \cite{Toulmin58}, \cite{WatzlawickWeaklandFisch74}.

Our proposition is instead pretty technical and formal. Operational
Research and Decision Analysis are seen as part of a more general
Decision Aiding Methodology (see \cite{Tsoukias08ejor}) aiming to
help real decision makers to understand, formulate and model their
problems and possibly reach a reasonable solution (if any). We are
concerned by that type of activities occurring in a decision aiding
situation where a ``client'' (very broadly defined) asks for some
advice or help to an ``analyst'', such an advice being expected to
come under form of a formal model allowing some form of rationality (see \cite{MeinardTsoukiasEJOR2018}). We call such activities a ``decision aiding process'' (see \cite{Tsoukias07aor}). At a certain point of that process the analyst will have to formulate a ``decision problem'' requiring some
computing to be performed by some algorithms, providing a result which
is expected to be used in order to present a recommendation relevant
to the decision maker's ``decision problem''.

Our focus is exactly here: what is a decision problem  for the analyst? The
proposal of the paper is to suggest a general framework within which it is
possible to identify all possible models, algorithms, procedures which
routinely analysts use in their job as well as to allow to invent ones (if
possible). General frameworks have been suggested in the literature since the existence of Operational Research and Decision Analysis (see \cite{CharnesCooper61}, \cite{French88}, \cite{LuceRaiffa57}, \cite{Vajda56}). The fact is that all such frameworks are based upon classes of modelling options or solution methods (mathematical programming, utility theory, multiple criteria decision analysis, game theory) and are constrained by the algorithms to be used in order to compute solutions. This is visible to all textbooks and manuals in our field (see for instance \cite{EhrgottFigueiraGreco2005}, \cite{Ravindran2008}) as well as consulting the list of keywords for the major OR conferences or journals: these include either methods or application fields. There is no categorisation independent from the solving methods and algorithms.

The framework we suggest aims instead at introducing a structure which allows
to categorise problems without any reference on which method is going to be
used to solve them. ``Methods'' should then follow as a consequence of such
categories and not vice-versa. Actually we may consider that decision problems could require the use of several different methods in order to handle their complexity. Such a challenge needs to address a number of difficult questions which we partially consider in this paper.

\begin{itemize}
 \item Which are the necessary and sufficient features in order to
     characterise any potential decision problem?
 \item Which is the strictly necessary information the client needs to
     provide in order to allow the analyst to model a decision problem?
 \item Which are the strictly necessary questions the analyst needs to make
     to the client in order to make a meaningful decision model?
 \item Which are the conditions of validity to be met in order to be
     confident about the model and the recommendations could entail?
\end{itemize}

We do not claim this paper provides an answer to all the above questions. We
claim instead we have some important intermediate results. We established the
necessary features describing the whole space of decision problems as well
the primitives (the strictly necessary information) in order to handle such
problems.

Practically speaking such results imply that the possible archetypes of
decision problems is finite (given that the number of features characterising
them is finite and discrete). They also imply that the classes of archetypal methods, depending upon the possible combinations of the primitives, are also finite.

The paper is organised as follows. Section 2, introduces the fundamental concepts and the notation used all along the paper. Then Section 3, introduces and explains the notion of ``primitives'' which is central for the construction of our framework: the minimal necessary information allowing to handle a decision problem. Section 4, analyses how decision support methods are designed focussing on two notions: preference aggregation and optimisation. Section 5, presents two examples which allow to understand why our framework is both interesting and generic. We discuss our findings in Section 6, before concluding.

\section{Concepts and Notation}

In order to be as precise as possible we are going to establish
a-priori an interpretation of a number of concepts which we are
going to use in the rest of the paper. We also introduce some
notation.

\begin{itemize}
 \item A procedure is a sequence of elementary mathematical operations
     transforming a mathematical structure to another one. Typical examples
     include: inverting a matrix, multiplying two vectors, ordering two
     numbers, transitively closing a graph, closing a logic formula,
     computing an average etc..
 \item An algorithm is a sequence of procedures with a precise information
     manipulation purpose. Typical examples include: the simplex algorithm,
     the variants of the SAT-algorithm, natural deduction, finding a kernel
     in a graph, linear regression etc..
 \item A protocol is a sequence of procedures and/or input/output steps
     allowing to collect information and to interact with the client. Typical examples are the construction of an utility function, indifference swaps in conjoint measurement, encoding beliefs etc..
 \item A method is a set of algorithms and protocols allowing to elaborate
     the information provided by the client for some decision aiding purpose. Typical examples include: the Branch and Bound method, multiple criteria decision analysis methods, preference learning methods etc..
 \item A model is a structural representation of the information elaborated
     in a decision aiding process and for the time being we are going to use two types of models: the ``problem formulations'' and  the ``evaluation models'' as introduced in \cite{Thebook00} and \cite{Tsoukias07aor}.
\end{itemize}

Recall that we are considering a formal definition of what a decision problem is. A first intuitive remark is that for this purpose we need to start from a set (representing feasible solutions of whatever the problem is) upon which we are going to construct our recommendation. We agree to denote this set as $A$ and for the time being we accept any type of element within it (including the empty set consisting in doing nothing). However, most of the times the set $A$ is more than just an enumeration: it will typically be described against different dimensions/attributes/variables/points of view, a set we will denote as $D$. A second remark is that in order to establish more formally our problem we need to define what type transformations the set $A$ is going to undertake: we will denote this as a ``problem statement'' $\Pi$. We will denote the collection of these three concepts as a problem formulation $\Gamma=\langle A,D,\Pi\rangle$. More details about these concepts are going to be discussed in the next Section.

We are now able to introduce the first definition of our paper.

\begin{deff} \label{deff1} A decision problem for the analyst consists in finding an appropriate partitioning of the set $A$, relevant for the decision maker's concerns. \end{deff}

\begin{remark}
  The reader should note that the definition applies also in the case the
  partitioning is ``uncertain'': such as fuzzy partitions (where elements of
  $A$ could have variable membership to different equivalence classes) or
  interval partitions (where elements of $A$ could belong to the union of
  equivalence classes).
\end{remark}

Along the paper we are going to use extensively preference relations. The basic relation we will adopt will be $\succeq$ which will read as ``at least as good as'' ($\succ$ will represent the asymmetric part of $\succeq$, while $\sim$ will represent the symmetric part). We will only make the hypothesis that this is a reflexive binary relation. The interested reader can see more about properties, preference structures and related matters in
\cite{OzturkTsoukiasVincke05} or \cite{RoubensVincke85} from which we
adopt definitions and notation.

\begin{figure}[h]
 \begin{center}
  \begin{picture}(250,60)\thicklines
   \put(5,45){\makebox(0,0){Ground}}
   \put(5,35){\makebox(0,0){Information}}
   \put(45,40){\vector(1,0){50}}
   \put(130,40){\makebox(0,0){Primitives}}
   \put(160,40){\vector(1,0){50}}
   \put(235,40){\makebox(0,0){Input}}
   \put(70,20){\vector(0,1){20}}
   \put(70,15){\makebox(0,0){Learning}}
   \put(70,5){\makebox(0,0){Protocols}}
   \put(185,20){\vector(0,1){20}}
   \put(185,15){\makebox(0,0){Modelling}}
   \put(185,5){\makebox(0,0){Tools}}
  \end{picture}
 \end{center}
\caption{The modelling process} \label{dialogue}
\end{figure}
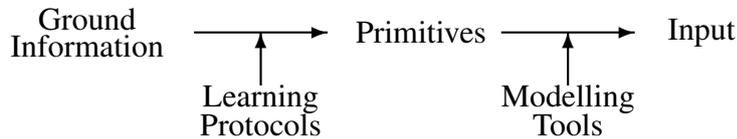

In this paper we are going to adopt a scheme of the modelling process as
depicted in Figure \ref{dialogue}. The idea is to separate the raw information as provided by the client (Ground Information) from the necessary information to build a decision support model (Primitives) and this from the information provided to an algorithm used for decision support purposes (Input).


\noindent More precisely we have: \\
\emph{Ground Information} consisting in collecting the client's statements about the problem for which she asked an advice. These come under the form of dimensions which matter for the decision (and these can be values, opinions or scenarios), elements of how a solution can be designed and preference statements.

\emph{Learning Protocols} are procedures allowing to identify preference
statements within the client's discourse and to translate them in ordering
relations. In order to do so we need to establish (in agreement with the
client) the sets on which such relations apply.

\emph{Primitives} is the strictly necessary formalised information the client needs to provide to the analyst in order to engage the decision aiding process. As noted these can be provided either directly (ground information) or indirectly (through learning protocols).

\emph{Modelling Tools} are the usual analytic tools an analyst uses in order to transform primitives in decision aiding objects. Examples include the
procedures allowing to construct a value function, a set of constraints, a
probability distribution etc..

The \emph{Input} is the information modelled in such a way that a decision
aiding method can be applied.

The reader should retain that the modelling process is not straightforward. The ground information usually comes in a ``messy'' form and it takes some effort to learn and construct the primitives out of it. It should also be noted that are the primitives that constraint the use of certain methods. If the primitives do not allow to construct a probability distribution, methods modelling uncertainty under such a form are unsuitable and we need to look for alternative formalisms. Last, but not least, it should be clear that
the primitives represent a commitment for the client. The analyst should make clear to the client that without establishing such primitives it is impossible to provide any reasonable support (at least in a formal way). A client can be evanescent, but up to a certain limit.

\begin{example} \label{gothamcity}
  \underline{\emph{Ground Information}}. Consider a city of $N$ districts of which we know the geography, the network and the distances. A client claims she needs to cover the city with a number of facilities. She asks an advice on where these should be located in order to minimise the openings. \\
  \underline{\emph{Learning Protocols}}. A discussion with the client, besides an analysis of her claims reveals that the two dimensions which matter for her decision is the covering of the city and the number of openings. Moreover, the notion of coverage is established on the hypothesis that an opening to any district ``covers'' the ``adjacent'' districts. \\
  \underline{\emph{Primitives}}. We call ``opening'' the decision to locate a facility to one among the 20 districts of the city. We denote with $x_j\in\{0,1\}$ the location decision with $j=1\cdots N$. The set $A$ of potential solutions is then the combinatorial structure $A=\{0,1\}^{20}$. Consider two such solutions $\chi,\psi\in A$ and two distinct dimensions: covering and opening. We establish two preference relations: $\succeq_c\subseteq A\times A$ and $\succeq_o\subseteq A\times A$ such that $\chi\succeq_c\psi$ iff the covering provided by $\chi$ is not inferior to the covering provided by $\psi$ ($c(\chi)\geq c(\psi)$) and $\chi\succeq_o\psi$ iff the openings foreseen by $\chi$ are not superior to the opening foresee for $\psi$ ($o(\chi)\leq o(\psi)$). \\
  \underline{\emph{Modelling Tools}}. In order to be more operational we need to give a more quantitative and formal model for the two dimensions of opening and covering. As far as opening is concerned we establish a function $O:A\mapsto\{0,20\}$ such that $o(\chi)=\sum_{x_j\in\chi}x_j$ (the opening of a solution being the number of districts in which a facility is located). As far as the covering is concerned we first introduce an ``adjacency relation'' $\gamma\subseteq J\times J$ ($J$ being the set of district indices), establishing an adjacency matrix $\mathcal{G}$. Given a district $j$ we denote with $\mathcal{G}(j)=\{i:\gamma(i,j)=1\}$ (the set of districts being adjacent to $i$). Given a solution $\chi$ the covering of $\chi$ will be a function $C:A\mapsto \{0,20\}$ established as $c(\chi)=|\{i:\forall x_j\in\chi\;x_j=1\wedge\gamma(i,j)=1\}|$ (the number of districts covered by any opening in $\chi$). \\
  \underline{\emph{Input}}. Since the client claims that any solution needs to ``cover'' the city, the covering dimension will only discriminate between acceptable (the ones satisfying full coverage) and unacceptable solutions. However, among all such acceptable solutions we seek the ones which minimise the openings. The result will be a well known combinatorial optimisation problem:

  \begin{eqnarray}
   \nonumber 
    ~ & \min\sum_jx_j & ~ \\
    \forall i\in J & \sum\gamma(i,j)x_j\geq 1 & ~ \\
    ~ & x_j\in\{0,1\} & ~
  \end{eqnarray}
  \hfill $\Box$
\end{example}

%
\section{Primitives}\label{primitives}

Let's analyse now in more details and more formally the concept of primitives. As already introduced, the idea is to identify the strictly necessary information with which we can construct a reasonable advice to the client. We can summarise this information in three topics: \emph{the problem statement}, \emph{the set upon which the preferences are expressed and the attributes or dimensions used to express preferences} (these three establishing the problem formulation), plus \emph{the preferences themselves}. We will present these topics in that precise order for reasons which will be clear at the end of the section.

\subsection{Problem Statement}{\label{Pi}

The notion of problem statement has already been introduced in
\cite{Bouyssouetal2006,ColorniTsoukias2013,Tsoukias07aor}. As already introduced in Definition \ref{deff1} any decision problem we may consider (both in theory and in practice) boils down to constructing a set $A$ and then partitioning it in $n$ classes. The concept of partition is well defined and considering (from a formal point of view) a decision as the partitioning of a set is simple and robust: it applies equally well in presence of fuzzy sets (fuzzy partitions) and in presence of different forms of uncertainty, including the case where an element of the set $A$ could belong to a union of several equivalence classes (of the set $A$) without knowing which one precisely.

There are several ways in which the partitioning of $A$ can be done, resulting in different ways to construct the equivalence classes $[A]_i$ ($i = 1, \ldots, n$) to which the set $A$ is partitioned. We focus on two different types of choices.

\begin{enumerate}
  \item The first choice concerns the ``comparisons'' used in order to construct the classes of the partition. Such preferences come under the form of a binary relation denoted by $S$ and for which we have two options:
      \begin{itemize}
       \item $S$ is a relation defined on $A$, in other terms $S\subseteq
           A\times A$, and we call this a ``relative comparison'';
       \item $S$ is a relation comparing the elements of $A$ to some
           externally defined norms (or standards) and vice versa. More
           precisely, if $N$ is the set of norms, then $S\subseteq A\times N\cup N\times A$, and we call this an ``absolute comparison''.
      \end{itemize}
  \item The second choice concerns the existence or not of an ``order'' upon the classes of the constructed partition. Let us make this point
      clearer. Consider $[A]=\{ [A]_1, \ldots, [A]_n \}$ the set of all classes partitioning $A$. Let us define a binary relation $\succcurlyeq \, \subseteq [A] \times [A]$ of the type ``at least as good as'', separable in a symmetric part ($\sim$) and an asymmetric ($\succ$) part. We have two cases:
      \begin{itemize}
       \item $\succcurlyeq=\emptyset$ which amounts saying there is no
           order among the  classes;
       \item $\succcurlyeq$ is reflexive, antisymmetric and transitive (a partial order), with a nonempty asymmetric part $\succ$. In this case, we consider the classes as (at least partially) ordered. The reader will note that this option is compatible with stronger ordered structures such as complete orders or even complete orders admitting a numerical representation on an interval scale or ratio scale. However, such strong ordered structures are not necessary.
      \end{itemize}
\end{enumerate}

Combining the above two choices, we can define four basic problem statements:
\begin{itemize}
 \item[--] ranking (partitioning  in ordered  classes, using relative comparisons);
 \item[--] rating (partitioning  in ordered  classes, using absolute comparisons);
 \item[--] clustering (partitioning in unordered  classes, using relative comparisons);
 \item[--] assignment (partitioning in unordered  classes, using absolute comparisons).
 \end{itemize}
For all the above cases, there are at least two special subcases, relevant from an algorithmic point of view (because result in specific procedures fitting these specific requirements):
\begin{itemize}
 \item[--] when the  classes are just two (complementary);
 \item[--] when one or more  classes have a given cardinality.
\end{itemize}

For instance, a ``choice'' problem statement is  a ranking problem statement
with only two classes: the elements to be chosen and the rest. The reader will note that most optimisation problems fit the ranking problem statement with only two  classes (the chosen/optimal ones and the rest). Equally easy is to realise that most pattern recognition, diagnosis signal processing, and
classification problems fit the assignment problem statement, while most data
analysis problems correspond to either clustering or rating problem statements.

We can now state our first fundamental claim upon which we build our theory.

\begin{claim} \label{claim1}
  Any decision problem belongs to one of the four categories: ranking, rating, clustering and assignment.
\end{claim}

\begin{remark}
  We draw the attention of the reader to the fact that a ``real decision aiding process'' typically consists in solving a sequence of formal decision problems of the type we defined in this paper. Each of such decision problems may be characterised by a different problem statement. Under such a perspective decision aiding consists in handling mix of problem statements.
\end{remark}

\subsection{Attributes and sets}\label{attrib}

Let's start talking about the set $A$ and how this is constructed in more details. The reader should note that our task is not to have a complete description of what the alternative options of action could be for the client, but a description which matters for her/him. The presentation here follows essentially the text appeared in \cite{ColorniTsoukias2020}. We consider two different spaces within which the set $A$ can be described: the variables space and the values space. In the first case we consider that any element of $A$ results from the combination of ``elementary choices'', instantiating variables from different domains. In the second case we consider that each element of $A$ has an image in a space of ``descriptors'' or ``features'' describing specific potential consequences. In other terms we consider the set $A$ as possible realisations of the variables $x_i$, each realisation having an image in a space of valuable consequences. More formally:

\begin{enumerate}
  \item The variables space is the product space of all the variables which
      might be used in order to compose an alternative. We consider $n$
      independent variables $x_i\;\;i\in\{1\cdots n\}$. If $\forall
      i\;x_i\in\mathds{R}$ then $A$ is a subset of a vector space
      ($A\subseteq\mathds{R}^n$). If $\forall i\;x_i\in\mathds{Z}$ then $A$
      is a combinatorial structure ($A\subseteq\mathds{Z}^n$). Without loss
      of generality we will only consider the case where $\mathds{Z}=\{0,1\}$ since
      any other combinatorial structure can be obtained from that one. There is of course always the case where the set $A$ is an
      enumeration of ``objects'' (a list). In this case we will consider that each alternative corresponds to a single variable. If we call the domain of each variable $x_j$ as $X_j$, then clearly $A=\prod_jX_j$. Hereafter we will represent a generic element of $A$ as $<x>$.
       \begin{example} Consider a well known production management problem whe\-re the system (under consideration) produces a number of different items $x_1,x_2,$ $\cdots,x_n$. If for each item we have a continuous set of feasible values $X_j$ then a solution (en element of $A$) is a bundle $\langle \bar{x_1},\bar{x_2},\cdots,\bar{x_n}\rangle$ where $\bar{x_j}\in X_j$ is an instance. Consider instead a the case of a set of candidates for a scholarship. Then each candidate will be represented by a variable $x_j$ whose domain $X_j=\{0,1\}$ (being chosen or not). The set $A$ will be any feasible combination of $x_j$ (for instance if only one scholarship is available then only combinations of cardinality 1 are feasible). \hfill $\Box$ \end{example}
  \item The values space, where each element of $A$ (independently from how it has been composed in the variables space) is mapped, is a set of attributes or evaluation dimensions $D$. Each $D_j\;(j\in\{1\cdots m\})$ should be seen as a function $D_j:A\mapsto E_j$ where $E_j$ is the set of all possible values an object can take under that attribute (often the domain $E_j$ is called the scale of the attribute $D_j$). Of course each $E_j$ can be either an interval of the reals or of the integers. There is however, a special case where the domain $E_j$ is composed by nominal labels (for instance colours: $E_j=\{$red,yellow,green...$\}$). Without loss of generality we will associate to this set a set of integers, paying attention not to consider the underlying ordering structure of the numbers. Independently of how the domains $E_j$ are established we denote the set $D(A)\subseteq\prod_jE_j$ as the image of $A$ in the values space.
       \begin{example} Consider a set of lottery tickets $\{a,b,c,\cdots\}$. To each ticket we associate a set of possible outcomes depending on whether the ticket is extracted as first prize, second prize, or no prize at all. We can denote this through scenarios $0,1,2,\cdots$ the scenario $0$ representing not being extracted. We obtain for each ticket a vector $\langle a_0,a_1,a_2\cdots\rangle$, $\langle b_0,b_1,b_2\cdots\rangle$ which represent all possible payoffs for each ticket. Equally, considering the production system case if a solution is a bundle $\langle \bar{x_1},\bar{x_2},\cdots,\bar{x_n}\rangle$, then given two attributes such as the price and the workforce consumption for any solution will be represented as $p(\langle \bar{x_1},\bar{x_2},\cdots,\bar{x_n}\rangle)$ or $w(\langle \bar{x_1},\bar{x_2},\cdots,\bar{x_n}\rangle)$. \hfill $\Box$ \end{example}
  \item There might be connections between the two spaces. More precisely it is often the case that each single variable (used in order to compose the set $A$) can be itself assessed against the attributes $D$. In other terms we may have that $D_j(\langle x_1\cdots x_n\rangle)\;=\; F(D_j(x_1)\cdots D_j(x_n))$, $F$ being an aggregation function, not necessarily linear.
       \begin{example} Consider again the production system case. If a solution is a bundle $\langle \bar{x_1},\bar{x_2},\cdots,\bar{x_n}\rangle$, and we can associate a price for each single variable $x_j$ then the price of any solution will be represented through a function $F(\langle p(\bar{x_1}),p(\bar{x_2}),\cdots,p(\bar{x_n})\rangle)$ possibly additive: $p(\langle \bar{x_1},\bar{x_2},\cdots,\bar{x_n}\rangle)=
       p(\bar{x_1})+p(\bar{x_2})+\cdots+p(\bar{x_n})$. \hfill $\Box$ \end{example}
  \item The important concept to bear in mind is that the set $D$ should
      satisfy ``separability''. There might be infinite different dimensions under which the elements of $A$ can be described and assessed. What we are interested are the ones which are relevant for the client. The way to check it is whether the client would use that precise dimension in order to discriminate two alternatives (for the rest identical) in case of a decision to be taken. If yes, it means that this dimension matters to the client and is separable, otherwise it remains a potential descriptor of $A$, for the moment irrelevant for the client and the decision process.
       \begin{example} Consider a set of patients who need to be sorted to different hospital departments. Although we may know several information about them (sex, age, residence, height, mass etc.), not all of them are relevant for being sorted within the hospital. Most of the times only the symptoms will be considered as relevant information for this assignment decision. The rest of the attributes are thus, non separable for this decision problem. \hfill $\Box$ \end{example}
\end{enumerate}

\begin{example}
  Let's discuss the above through a very well known example: the ``knapsack
  problem''. There are $n$ different objects, each of which can be chosen in
  order to be carried within a container (the knapsack). It turns natural to
  associate to each such object a variable $x_j\in\{0,1\}$ representing the
  choice to carry or not that specific object. The set $A=\{0,1\}^n$ is thus described in the variables space by the $2^n$ possible ``bundles'' of objects. Let's consider now three attributes: value, weight and colour to be used in order to assess the different possible bundles (alternatives). Let's now make the hypothesis that we know the value, weight and colour of each single object: we will represent them as $v(x_j)$, $w(x_j)$ and $c(x_j)$. It is natural to consider that $\forall <x>\;\in A\;w(<x>)=\sum_jw(x_j)$ (the weight of each bundle will result as the sum of the weight of the objects composing it). It is equally easy to understand that we cannot use such a linear aggregation as far as the colour is concerned. The value attribute could be linear ($\forall <x>\;\in A\;v(<x>)=\sum_jv(x_j)$, but only if there are not ``absorbing values among objects'' (if I pick $x_1$ is useless to pick $x_2$). Last, but not least, the discussion with the client could reveal that finally the colour is not a separable attribute, in the sense that she/he would not make any decision just because two bundles have a different colour.
\end{example}

%
We are now able to introduce our second fundamental claim.

\begin{claim} \label{claim2}
  A decision problem exists if there is at least one separable attribute
  describing the set $A$.
\end{claim}

Remember that a decision problem corresponds to a partitioning of the set $A$
(using one among the four fundamental problem statements). However, such a
partitioning can occur if there is at least one dimension, considered relevant by the client, to be used in order to discriminate the objects among them: a separable attribute. The issue however, is to understand where the set $A$ comes from. It can be certainly be provided by the client as an input, but most of the times this is not the case. Clients have a vague idea of what they have to decide: it is part of the decision aiding process to construct a precise set of alternatives upon which we can apply a formal decision problem. How is the set $A$ constructed?

We can now state the first result of our paper.

\begin{prop} \label{prop1} Constructing the set $A$ is itself a decision problem.
\end{prop}

\noindent\textbf{Proof}. Suppose a decision situation for which the client claims that
the set $A$ is totally unknown. If this is to be considered as a decision
problem then exists at least one separable attribute which is able to
discriminate and assess the (unknown) set $A$. We already know that for each
such separable attribute exists a set $E$ of values (representing the domain of the attribute). The simple hypothesis to do is $A=E$. In other terms we assume that exists a decision variable having as domain $E$. This automatically establishes a set $A_0$ upon which we can apply a partitioning procedure aiming at satisfying the client's requirements. There are two cases: \\
 1. The partition is satisfactory. The procedure stops, since the client is
 satisfied. \\
 2. The partition is unsatisfactory. We have again two, non exclusive, cases:
 \\
  2.1. Add further separable attributes, under which we can refine further the partitioning of $A_0$. \\
  2.2. Add further decision variables enriching the set $A_0$. \\
 In both cases we generate a new set $A_1$, using the partition of $A_0$ and
 the new attributes and/or variables introduced. We can now apply upon $A_1$ a new partitioning procedure. \\
Consider now the generic step $i$ of this procedure where
$A_i=\bigcup_i[A_{i-1}]_i$. If this partitioning is satisfactory then we end,
otherwise we repeat the procedure. This will end when the client declares to be satisfied. \hfill $\blacksquare$

%
\textbf{Discussion}. The case where the set $A$ is totally unknown is rather
unrealistic. In most of the cases the client comes with a vague idea of what
this set should look like and we are able to establish some decision variables describing the composition of $A$ and some separable attributes assessing it. However, most of the times this set will result to be ``unsatisfactory'' and the way through the client (and the analyst) realise it is that, whatever the partitions generated out of this set, the client is unsatisfied with the result.

Under such a perspective what the theorem practically tell us is that the
construction of the set $A$ is a process (part of the whole decision aiding
process) combining two activities: one, creative, consisting in identifying new attributes and/or variables and one, technical, consisting in solving a
decision problem generating new partitions of the incumbent set $A_i$ (at step $i$ of the process; the reader can check the similarities of this idea with the concepts of ``expansive partitions'' in formal design theory:
\cite{HatchuelWeil2009}). This process is subjectively driven by the client's
satisfaction (as always in a decision aiding process).

\subsection{The preferences}\label{preferences}

The interested reader can have more information on this topic in \cite{Fishburn70}, \cite{Morettietal2016}, \cite{Pigozzietal2016}, \cite{RoubensVincke85}. Let's start talking about ``preference statements''. These are sentences expressed (typically) by the client within which we can identify essentially two objects (implicitly or explicitly): \\
 - one or more sets upon which preferences are expressed; \\
 - one or more binary relations representing formally such preferences. \\
Let's give some examples.

\begin{example} \label{preference-cases}
  In the following $X,Y,Z\cdots$ represent elements of the set upon which preferences are expressed. \\
  1 $X$ is nice; \\
  2 $X$ fits better than $Y$; \\
  3 $X$ has a very bad score; \\
  4 $X$ is very similar to $Y$; \\
  5 $X$ meets the requirements; \\
  6 $X$ and $Y$ will do better than $X$ and $Z$; \\
  7 $X$ is preferred to $Y$ more than $Z$ being preferred to $W$; \\
  8 A ``red and long stick'' is better than a ''yellow and short'' one. \\
  9 Price is more important than comfort.
\end{example}

What these sentences reveal? First of all there are three different sets that are involved: the first one is the set of objects (we call it $A$) which is assessed, then occasionally a set of norms or standards (we call it $N$; when se say that $X$ is ``nice'' we implicitly assume that somewhere exists a standard of ``nice'' to which $X$ is compared), and then there is a set ($D$) of attributes upon which a preference relation may be expressed.

Then we have the ordering or preference relations: in the following we will use a generic binary relation $\succeq$, which is reflexive and decomposable in an asymmetric part ($\succ$) and a symmetric one ($\sim$); any of these being potentially empty. The reader should note that we use the term ``preference'' also in the case we consider only symmetric comparisons such as similarities. We get the following types of preference statements: \\
 - if $\succeq\subseteq A\times A$ then we call $\succeq$ a first order relative comparison; \\
 - if $\succeq\subseteq A\times N\vee N\times A$ then we call $\succeq$ a first order absolute comparison; \\
 - if $\succeq\subseteq 2^A\times 2^A$ then we call $\succeq$ a first order extended relative comparison; \\
 - if $\succeq\subseteq A^2\times A^2$ then we call $\succeq$ a first order preference intensity comparison; \\
 - if $|D|\geq 2$ and $\succeq\subseteq D(A)\times D(A)$ then we call $\succeq$ a first order multi-attribute relative comparison; \\
 - if $\succeq\subseteq D\times D$ then we call $\succeq$ a first order relative importance or a second order relative comparison (and we denote it $\gg$); \\
 - combinations of the above (for instance we can have first order multi-attribute preference intensity comparisons or second order extended relative comparisons etc.).

\begin{example}
  Consider the cases presented in example \ref{preference-cases}. Case 1 is a First Order Absolute Comparison, case 2 is a First Order Relative Comparison, case 3 is a First Order Absolute Comparison, case 4 is First Order Relative Comparison, case 5 is a First Order Absolute Comparison, case 6 is a First Order Extended Relative Comparison, case 7 is a First Order Intensity Relative Comparison, case 8 is a First Order Multi-Attribute Relative Comparison, case 9 is a Second Order Relative Comparison.
\end{example}

There are two more issues we need to address when preferences need to be modelled on different multiple dimensions for decision purposes. \\
 - The first one consists in checking preferential independence. Consider two preference statements of the type $x\succeq_1 y$ and $x\succeq_2 y$. If for any reason there is a logical dependance between the two statements (of the type $\forall x,y \;\;\phi(x\succeq_1 y)\rightarrow\psi(x\succeq_2 y)$, where $\phi,\psi$ are logical connectives) we consider the preferences to be conditional. Otherwise we satisfy preferential independence. For more details on this topic the reader can see \cite{BoutilierBrafmanHoosPoole99}, \cite{Thebook00}, \cite{Wakker89}. \\
 - The second one consists in distinguishing explicitly the semantic difference between $x\succeq y$ and $\neg(x\succeq y$) when these two sentences are not one the complement of the other. In these cases we talk about explicit ``negative'' preference statements, a typical example being the use of ``veto conditions'' while stating a decision rule (see more in \cite{TsoukiasFCDS02}, \cite{TsPerVin02}, \cite{TsoukiasVinckeTHEO95}).

\subsection{More about the primitives}\label{Moreprim}

Let's summarise our concepts and findings. As already introduced the concept of a primitive for a decision problem is related to the strictly necessary information required in order to be able to elaborate a decision model and a recommendation for the client. We can now give a more formal definition.

\begin{deff} \label{deff2}
  Primitives are pieces of information relevant to a decision aiding process which can only be obtained from the ground information and the learning protocols to submit to the client.
\end{deff}

Under such a perspective the set $D$ of separable attributes is a primitive because we can only get it discussing with the client about what matters for her. The set $A$ instead, since can be constructed elaborating upon the set $D$ is not strictly speaking a primitive. However, in case we need to combine decision variables in order to construct elements of $A$, these should be considered as primitives since we can only get them from the client.

Certainly, preferences are primitives: we cannot really elaborate a decision model and any recommendation without knowing the client's preferences. However, not any type of preference statements are primitives. For this purpose we present two specific propositions.

\begin{prop}\label{secondnotprim}
  Second order preference statements are not primitives.
\end{prop}

\noindent \textbf{Proof}. In order to show the proposition we need to show that there is always a procedure through which we can construct a second order preference statement from other primitives or constructed information. Consider the set $D$ of dimensions and consider two subsets $H$ and $G$ of $D$, such that preferences expressed upon $H$ are independent from those expressed upon $G$. Then it is sufficient to set up a learning protocol identifying elements of $A$ such that $x\succeq_H y$, $y\succeq_G x$ and $x\succeq_{H\cup G}y$, which can only be justified by the existence of a binary relation upon $2^D$ such that $H\gg G$ (which is a second order preference statement). In case preferential independence does not hold then the existence of a logical implication among first order preference statements establishes a connection between subsets of dimensions and thus a second order preference statement. \hfill $\blacksquare$

\begin{prop} \label{intensnotprimit}
  First order preference intensity comparisons are not primitives.
\end{prop}

\noindent \textbf{Proof}. Once again it is sufficient to show the existence of a procedure through which we can construct preference intensity comparisons from other primitives. This is the case, considering the procedure of indifference swaps through which we construct value functions (which are actually a measure of preference intensity). \hfill $\blacksquare$

\vspace{5mm}

\textbf{Discussion}. What the propositions tells us is that what we strictly need during a decision aiding process are first order relative or absolute comparisons, but we do not need second order ones, of the type commonly known in the literature as relative importance parameters or ``weights'' (see \cite{podin1mss94}, \cite{BRVM96}). This is compatible with the existing literature (see \cite{Bouyssouetal2006}, \cite{Marchant03}) where the notion of ``relative importance'' is conditional to first order preference statements. Extending the notion of relative importance to likelihoods of scenarios (usually known as probabilities), the topic has already discussed independently in \cite{Ramsey31} and in \cite{deFinetti37} (see also \cite{Nau01}, since likelihoods can be derived comparing simple lotteries among them (first order preference statements).

The fact that certain information considered useful for the construction of a decision model and the conduction of a decision aiding process are not primitives does not mean these are useless or irrelevant. It simply tells us we need to assess whether it pays to obtain them directly or whether it should be preferred an indirect procedure constructing them. It also tells us that their absence is not a prejudice to a satisfactory conduction of the decision aiding process, although their presence can be of great help.

\section{Algorithms and Methods}\label{methods}

Once the notion of primitives is set up we need to understand how to proceed with constructing decision models and appropriate methods aimed at providing a ``solution'' to that decision problem and a recommendation to the client. Let us recall that in our setting we have introduced the notion of problem formulation as the triplet $\Gamma=\langle A,D,\Pi\rangle$, where $A$ is the set of alternatives assessed against $D$, the set of evaluation attributes (or dimensions) and $\Pi$ is the problem statement (how $A$ is partitioned).

Intuitively speaking, the different dimensions or attributes may have 3 different origins: \\
 - multiple values relevant for the decision and the client; \\
 - multiple opinions, of diverse stakeholders who might be relevant for the decision and the client; \\
 - multiple scenarios which may occur in the future and for which different preferences have to be expressed.

The generic situation therefore, is the one where preferences are expressed on any among these three types of dimension. We will represent this general framework in Figure \ref{cube}.

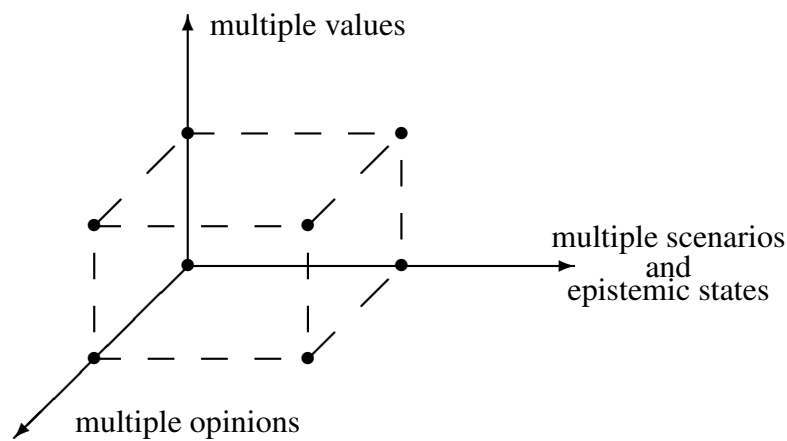
\begin{figure}[h]
 \begin{center}
  \begin{picture}(300,220)\thicklines
   \put(260,120){\makebox(0,0){multiple scenarios}}
   \put(260,110){\makebox(0,0){and}}
   \put(260,100){\makebox(0,0){epistemic states}}
   \put(125,200){\makebox(0,0){multiple values}}
   \put(80,50){\makebox(0,0){multiple opinions}}
   \put(80,110){\makebox(0,0){$\bullet$}}
   \put(160,110){\makebox(0,0){$\bullet$}}
   \put(160,160){\makebox(0,0){$\bullet$}}
   \put(80,160){\makebox(0,0){$\bullet$}}
   \put(45,75){\makebox(0,0){$\bullet$}}
   \put(45,125){\makebox(0,0){$\bullet$}}
   \put(125,75){\makebox(0,0){$\bullet$}}
   \put(125,125){\makebox(0,0){$\bullet$}}
   \put(80,110){\vector(1,0){145}}
   \put(80,110){\vector(0,1){95}}
   \put(80,110){\vector(-1,-1){65}}
   \put(125,125){\line(1,1){10}}
   \put(145,145){\line(1,1){10}}
   \put(125,75){\line(1,1){10}}
   \put(145,95){\line(1,1){10}}
   \put(45,125){\line(1,1){10}}
   \put(65,145){\line(1,1){10}}
   \put(45,125){\line(1,0){10}}
   \put(65,125){\line(1,0){10}}
   \put(85,125){\line(1,0){10}}
   \put(105,125){\line(1,0){10}}
   \put(45,75){\line(1,0){10}}
   \put(65,75){\line(1,0){10}}
   \put(85,75){\line(1,0){10}}
   \put(105,75){\line(1,0){10}}
   \put(80,160){\line(1,0){10}}
   \put(100,160){\line(1,0){10}}
   \put(120,160){\line(1,0){10}}
   \put(140,160){\line(1,0){10}}
   \put(45,85){\line(0,1){10}}
   \put(45,105){\line(0,1){10}}
   \put(125,85){\line(0,1){10}}
   \put(125,105){\line(0,1){10}}
   \put(160,120){\line(0,1){10}}
   \put(160,140){\line(0,1){10}}
  \end{picture}
 \end{center}
   \caption{Multiple Dimensions}\label{cube}
\end{figure}

It is easy to realise that there might be combinations of such dimensions. The opinion of a stakeholder can depend from multiple values and/or multiple scenarios she may want to consider. At the same time a value could be formed out of the opinion of different experts. It is equally easy to realise that the ``cube'' we show in Figure \ref{cube} is just a convenient way to represent how these three types of dimensions combine. In reality nothing impedes that this ``cube'' has more than 3 dimensions in case values, opinions and scenarios result from the combination of other values, opinions and scenarios. The intuitive idea here is that in order to compute a recommendation (or to solve a decision problem) we need to move along the edges of the ``cube'' in order to simplify the problem and aggregate the preferences when these are expressed on multiple dimensions at the same time. Said in other terms: if we know preferences expressed on several values (opinions or scenarios), considering that these will not be unanimous, we need to find a way to put them together, constructing a simpler primitive on a single dimension. Moving along the edges of the ``cube'' in Figure \ref{cube} represents exactly this type of simplification. This is essentially the idea of optimisation.

\subsection{Optimisation}\label{ssoptimisation}

In order to develop further our reasoning we need to introduce the concept of ``optimisation procedure''.

\begin{deff}\label{defopt}
  We define as optimisation procedure any algorithm solving the problem $\inf_{x\in A}f(x)$ where $A$ is the set of alternatives and $f(x)$ is some function from $A$ to a partially ordered set.
\end{deff}

The reader will note that we use the notation $\inf$ and not $\min$ since the image of $A$ for $f(x)$ might not be totally ordered. We can now state a new proposition for our theory.

\begin{prop}\label{ppoptimisation}
  If primitives (preferences) are expressed on a single dimension then any problem statement results in using an optimisation procedure for  partitioning $A$.
\end{prop}

\noindent\textbf{Proof}. Let's recall that we have a set $A$ and as a primitive a binary relation $\succeq\subseteq A\times A$ (or $\succeq\subseteq A\times N\cup N\times A$). Consider the problem statements where the equivalence classes are ordered (ranking and rating). Then given the relation $\succeq$ (which is not necessarily an ordering relation) we construct a binary relation $\succcurlyeq$ ``as near as possible'' to $\succeq$ and then extracting the $\inf$ of $\succcurlyeq$ is straightforward. We then repeat the procedure with the remaining elements of the set $A$. \\
Consider now the assignment problem statement. Assigning an element $x\in A$ to any among the predefined classes is equivalent to a constraint satisfaction problem, given than any logical clause we may conceive for the assignment rule can be translated to one or more constraints. But then any constraint satisfaction problem can be solved through an optimisation procedure (although not necessary; \cite{BisMR97}, \cite{BrailsfordPottsSmith99}, \cite{Tsang93}). \\
Finally, consider the clustering problem statement. Typically we will use the primitive binary relation to construct a ``distance'' (see \cite{Janowitz2010}, \cite{MeyerOlteanu2013}) and then optimise a ``fitting function''. \hfill $\blacksquare$.

\vspace{5mm}

\noindent\textbf{Discussion}. There are three remarks we need to do here. The first one concerns the fact that optimisation applies once we have a single dimension primitive upon the set $A$. This means we need to reduce our decision problem to a single dimension one. The second one concerns the fact that although any problem statement can be reduced at using an optimisation procedure, this does not imply that we always adopt an optimisation procedure in order to solve a precise problem. It often happens that other ad-hoc designed procedures might be more efficient. This does not invalidate our proposition: any type of decision problem, when primitives are on a single dimension is solvable through an optimisation procedure. The third remark is more general: the fact that in order to solve decision problems we use optimisation procedures does not mean that decision support, from an operational point of view, is optimisation. As it will be clear at the end of the paper (and as reported in the literature; \cite{Tsoukias07aor}) aiding to decide is a much more complicated process than solving one or more formal decision problems.

%
\subsection{Preference Aggregation}\label{aggregation}

Let's come back to Figure \ref{cube}. As already mentioned the general situation is the presence of multiple dimensions (under form of values, opinions, scenarios or combinations of these) through which preferences are expressed. If this is the case then the ``cube'' can be seen as a long hierarchy where a set of ``children nodes'' contributes defining the primitive over a ``parent node'', which combined with other nodes at this level contributes defining the primitive over a parent mode at a superior level etc., as shown in Figure \ref{tree}.

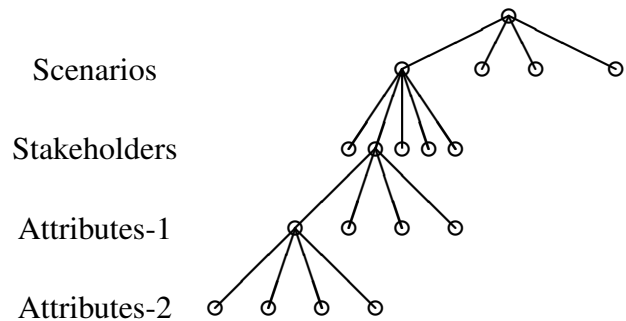
\begin{figure}[h]

\begin{center}
\begin{picture}(200,130)\thicklines
  \put(160,120){\circle{5}}
  \put(120,100){\circle{5}}
  \put(150,100){\circle{5}}
  \put(170,100){\circle{5}}
  \put(200,100){\circle{5}}
  \put(5,100){\makebox(0,0){Scenarios}}
  \put(100,70){\circle{5}}
  \put(110,70){\circle{5}}
  \put(120,70){\circle{5}}
  \put(130,70){\circle{5}}
  \put(140,70){\circle{5}}
  \put(5,70){\makebox(0,0){Stakeholders}}
  \put(80,40){\circle{5}}
  \put(100,40){\circle{5}}
  \put(120,40){\circle{5}}
  \put(140,40){\circle{5}}
  \put(5,40){\makebox(0,0){Attributes-1}}
  \put(50,10){\circle{5}}
  \put(70,10){\circle{5}}
  \put(90,10){\circle{5}}
  \put(110,10){\circle{5}}
  \put(5,10){\makebox(0,0){Attributes-2}}
  \put(160,120){\line(-2,-1){40}}
  \put(160,120){\line(-1,-2){10}}
  \put(160,120){\line(1,-2){10}}
  \put(160,120){\line(2,-1){40}}
  \put(120,100){\line(-2,-3){20}}
  \put(120,100){\line(-1,-3){10}}
  \put(120,100){\line(0,-3){30}}
  \put(120,100){\line(1,-3){10}}
  \put(120,100){\line(2,-3){20}}
  \put(110,70){\line(-1,-1){30}}
  \put(110,70){\line(-1,-3){10}}
  \put(110,70){\line(1,-3){10}}
  \put(110,70){\line(1,-1){30}}
  \put(80,40){\line(-1,-1){30}}
  \put(80,40){\line(-1,-3){10}}
  \put(80,40){\line(1,-3){10}}
  \put(80,40){\line(1,-1){30}}
\end{picture}
\end{center}
\caption{Many decision problems hierarchically related}\label{tree}
\end{figure}

\begin{remark}
  The reader should note that: \\
   - The hierarchy shown in Figure \ref{tree}, Scenarios, Stakeholders, Attributes-1, Attributes-2 is just an example. Any sequence of different type of dimensions could fit. \\
   - There might be cases where, despite the presence of different levels of such hierarchies (for instance in Figure \ref{tree} the bottom level is Attribute-2, then Attribute-1, then Stakeholders), the aggregation step might have to move two or more steps all-together because of preferential dependencies (back to  the example we might have to aggregate the two levels of attributes in one step although semantically different).
\end{remark}

In other terms, technically speaking, before we can optimise along a certain dimension (as from proposition \ref{ppoptimisation}), we need to aggregate the preferences (primitives) as these are expressed at the level of the nodes which contribute defining this dimension (if any). We need to handle a preference aggregation procedure \cite{Thebook00}, \cite{Bouyssouetal2006}. The reader should note that it does not matter if we aggregate preferences expressed under values, opinions or scenari. From a technical point of view the problem is the same: construct a binary relation upon the set $A$ out of a set of binary relations upon the same set $A$. We are not going to present any precise method for this purpose (the topic is extremely large and diversified; see \cite{Brandtetal2016book}, \cite{Regenwetter2009}, \cite{Saari1994}, \cite{Vincke82a}). We are going instead to present and discuss what we need to check before we can proceed with any preference aggregation procedure.

\begin{enumerate}
  \item The first issue we need to address is whether it is possible to use more than a simple binary relation when modelling preferences. More precisely, whether it is possible to measure the difference among preferences: know that the difference of preference between $x$ and $y$ is more than the difference of preference between $z$ and $w$. If this is the case we can construct measurement scales which allow to quantify proportions and distances of values. This is typically what happens when we use cost functions or value functions or any other measure of ``impact'' which is expected to provide a quantitative information (more than simply ordinal; see \cite{roberts79}). Actually this is the implicit hypothesis in almost all mathematical programming methods, models for decision under risk, quantitative economics etc..
  \item The second issue is related to commensurability. Supposing on each single dimension to be able to construct measurement scales allowing to compare differences of preferences, we need to check whether these are comparable among them. The question here is to know if the difference of preference between $x$ and $y$ on dimension $k$ is larger (or smaller) than the difference of preference between $z$ and $w$ on dimension $l$. If it is the case (and is not always the case), then we can construct appropriate ``trade-offs'' among the different dimensions: practically we can reduce all different dimensions to a single measurement scale, making extremely easy to aggregate preferences.
  \item The third issue has already been introduced in section \ref{preferences} and concerns preferential independence. Supposing we can measure differences of preferences and these are commensurable among the different dimensions, then linear models of aggregation require preferential independence among any subset of dimensions. Alternatively non-linear models might be deployed in order to take into account interactions among preferences expressed along the different dimensions. This holds for any type of preference aggregation procedure (see \cite{BoutilierBrafmanHoosPoole99}, \cite{14-MErgJFigSGre2016}).
  \item The fourth issue has also been introduced in section \ref{preferences} and consists in checking if explicit negative preference primitives are admissible. In other terms we need to know if the models satisfying $x\succeq y$ are the complement of the those satisfying $\neg(x\succeq y)$. If it is not (and there are plenty of real world situations and of models where it is not; see \cite{OzturkTsoukias08}, \cite{Ozturk:2007}) then we need to design appropriate methods to take this modelling option into account.
  \item Last, but not least we need to establish the properties the method, the result and the recommendation need to satisfy. Besides satisfying requirements of meaningfulness (see \cite{NarensLuce1990}, \cite{roberts79}, \cite{Roberts1980}) we may need to enforce a number of characteristics of our results and the method used. We may need the method to be explicable or we may need the result to satisfy anonymity, non-manipulability or scalability (see social choice theory; \cite{Brandtetal2016book}).
\end{enumerate}

A first consequence of the above presentation is that the number of archetypal methods for preference aggregation is bounded by the number of combinations of the possible answers to these questions (these being finite). In other terms there is a finite number of possible archetypal preference aggregation methods (see \cite{vinck1theo92}).

\begin{prop}
  The number of archetypal methods solving decision problems is finite.
\end{prop}

\textbf{Proof.} A method solving a decision problem is a combination of primitives and preference aggregation procedures (plus optimisation). Since the number of primitives is finite and the possible preference aggregation methods are also finite, the number of archetypal methods is also finite. \hfill $\blacksquare$.

\vspace{5mm}

We can now summarise our presentation in Figure \ref{agregsqaure} where arcs represent the existence of a procedure allowing to obtain the information at the destination node from the information available at the origin node. Usually our primitives are the binary relations $\succeq_j$ we see at the top left of the Figure. These are essentially learned from the ``client'' either directly or indirectly. If appropriate conditions are fulfilled then these binary relations can be represented by functions $f_j$ (measures) either simply ordinal or more than ordinal (ratio or interval scales). It could be the other way also. We come to know directly the functions (we get measures from some reliable source) and from these we infer the preferences represented by the binary relations. In both cases what we are looking for is either a binary relation which represents all the binary relations together or a function which does the same aggregating the functions. It could be that we can infer the global function from the global binary relation and vice-versa (if appropriate conditions are met). Typically we either aggregate the binary relations to a global binary relation or the functions to a global function. However it might be that we can directly aggregate the binary relations to a global function as it happens for conjoint measurement functions \cite{Bouyssouetal2006}, \cite{BouyssouPirlot2009book}.

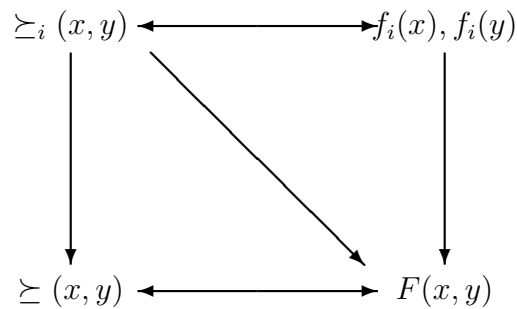
\begin{figure}[h]
\begin{center}
\begin{picture}(160,120)\thicklines
  \put(10,10){\makebox(0,0){$\succeq(x,y)$}}
  \put(10,110){\makebox(0,0){$\succeq_i(x,y)$}}
  \put(150,110){\makebox(0,0){$f_i(x),f_i(y)$}}
  \put(150,10){\makebox(0,0){$F(x,y)$}}
  \put(80,10){\vector(1,0){45}}
  \put(80,10){\vector(-1,0){45}}
  \put(80,110){\vector(1,0){45}}
  \put(80,110){\vector(-1,0){45}}
  \put(10,100){\vector(0,-1){80}}
  \put(150,100){\vector(0,-1){80}}
  \put(40,100){\vector(1,-1){80}}
\end{picture}
\end{center}
\caption{The preference aggregation problem}\label{agregsqaure}
\end{figure}

\section{Two examples}

It is time to start completing our journey. Let's make the point. For the time being we have shown that: \\
 - a decision problem is always handled through an optimisation procedure; \\
 - most of the times we need to aggregate preferences expressed under multiple dimensions (values, opinions or scenari) to one dimension primitive (where optimisation applies), a problem we call preference aggregation; \\
 - preference aggregation methods depend on how the five issues presented in section \ref{aggregation} are considered.

Recall that a decision problem essentially consists in partitioning a set (which we call $A$) and that we have already shown that constructing the set $A$ is itself a decision problem. We have shown that any method aimed at handling decision problems is a mix of three procedures: \\
 1. set construction procedures; \\
 2. preference aggregation procedures; \\
 3. optimisation procedures.

In order to show that, and at the same time explain our point of view, we are going to use two examples, one in combinatorial optimisation and the other in decision under risk, for which we certainly have well known methods to use, but for which our framework applies and can improve the whole decision aiding process.

\begin{example}
  We continue with example \ref{gothamcity}. Let's see now how our framework applies in solving this problem. At the beginning we have an initial set $A_0$ which is the whole combinatorial structure $\{0,1\}^{20}$ ($20$ binary relations resulting to $2^{20}$ possible combinations. How many evaluation dimensions do we consider? Since the client declares willing to ``cover'' the city, we need to ``cover'' each single district. We introduce n binary relations (one for each district $j$) of the type $\succeq_j\subseteq_j A\times\{0,1\}$ and we consider a rating problem for which, given any solution $\langle \bar{x}_1\cdots\bar{x}_n\rangle$, we want $\forall j\langle \bar{x}_1\cdots\bar{x}_n\rangle\succeq_j 1$. If we know the adjacency matrix $\cal{G}$, such that for any two given districts $i$ and $j$ $\gamma_{ij}=1,\;\gamma\in\cal{G}$ iff the two districts are adjacent we can rewrite the preference constraint in a more suitable form which is $\forall j\sum_{i}\gamma_{ij}x_i\geq 1$.

  The above define 20 rating decision problems (with two ordered classes: the feasible and the unfeasible solutions) which if solved one after the other (20 different steps of set construction) will result in a set $A_{20}$ which contains only feasible solutions for all 20 districts. We are now able to introduce a binary relation $\succeq_o\subseteq A_{20}\times A_{20}$ such that given two solutions (let's call them \textbf{1} and \textbf{2}) $\langle x^1_1\cdots x^1_n\rangle\succeq_o\langle x^2_1\cdots x^2_n\rangle$ iff $o(x^1_1\cdots x^1_n)\geq o(x^2_1\cdots x^2_n)$ where $o:A\mapsto\{1\cdots 20\}$ is a function measuring the openings. Clearly $o(x_1\cdots x_n)=\sum_jx_j$ and the ranking decision problem consists in solving $\min\sum_jx_j$. \hfill $\Box$.
\end{example}

\noindent\textbf{Discussion}. Many readers will naturally ask why do we need such a complicated version for a problem which is already well known in the literature. There are two reasons for which we believe such a presentation is convenient.

The first one is generality. What we want to show is that our framework applies to any type of decision problem and generalises all methods we can design in order to handle decision problems. Although this is a well known combinatorial optimisation problem we can apply to it the same process we may apply for any type of decision problem and which we summarise as follows: \\
 1. construct an initial set $A_0$ either as combination of descriptive attributes or as combination of values of evaluation attributes; \\
 2. identify the primitives (preferences on different several dimensions); \\
 3. aggregate preferences in order to create new primitives on less dimensions; \\
 4. refine the set $A_i$ (at generic step $i$); \\
 5. go ahead until the refined set is partitioned in a way which satisfies the client's demand.

The second one (related to the first one) is the possibility to revise and/or update the model in a clear and formal way as soon as there are reasons for which we need to do so. Since we have a general process we just need to go through it and make the appropriate modifications.

Suppose we want to relax the covering requirement: instead of covering the whole city (which might be too expensive) we just consider covering as much as possible. If this is the case, then covering becomes an evaluation dimension and for this purpose we also need to introduce for each district a new binary variable ($y_j$: being covered by a facility or not) besides the ``opening variables'' ($x_j$: open a facility or not). The result will be that our initial set $A_0$ will now be $\{0,1\}^{40}$ (we now have 40 binary variables). The 20 rating problems (equivalent to the feasibility constraints) will now be formulated as $\forall j\sum_i\gamma_{ij}x_i\geq y_j$. Further on we need to introduce a new preference relation $\succeq_c\subseteq A\times A$ such that (using the same notation of the example) $\langle x^1_1\cdots x^1_n\rangle\succeq_c\langle x^2_1\cdots x^2_n\rangle$ iff $c(x^1_1\cdots x^1_n)\geq c(x^2_1\cdots x^2_n)$ where $c:A\mapsto\{1\cdots 20\}$ is a function measuring the covering. Clearly $c(x_1\cdots x_n)=\sum_j\alpha_{ij}x_j$ and the ranking decision problem consists in solving $\max\sum_jy_j$ and $\min\sum_jx_j$.

Should we want to transform any of the two ranking problems to a rating problem (feasibility) we can introduce well known constraints of the type $\sum_jk_jx_j\leq K$ where $k_j$ is the cost of each opening and $K$ the available budget or $\sum_jp_jy_j\geq P$ where $p_j$ is the population of each district and $P$ the target to reach.

\begin{example}
 Alice is considering submitting a paper to a prestigious conference being very (very) selective (acceptance rate $r$). Besides, the conference location is not easy to reach and flying there can become very expensive. If Alice submits the paper ($s$) and buys the ticket now ($b$) it is still possible to find it at an affordable price ($t$). The problem is that she has to buy it using her own funds and she will be reimbursed only in case the paper is accepted. If she submits the paper and buys ($B$) the ticket only in case the paper is accepted the price will rise ($T$) and could exceed the department's budget ($K$), making impossible attending the conference (the probability of this being the case being $p$). Noting the reward for attending the conference with $R$, Alice's problem can be represented with the decision tree shown in figure \ref{decisiontree01}.

 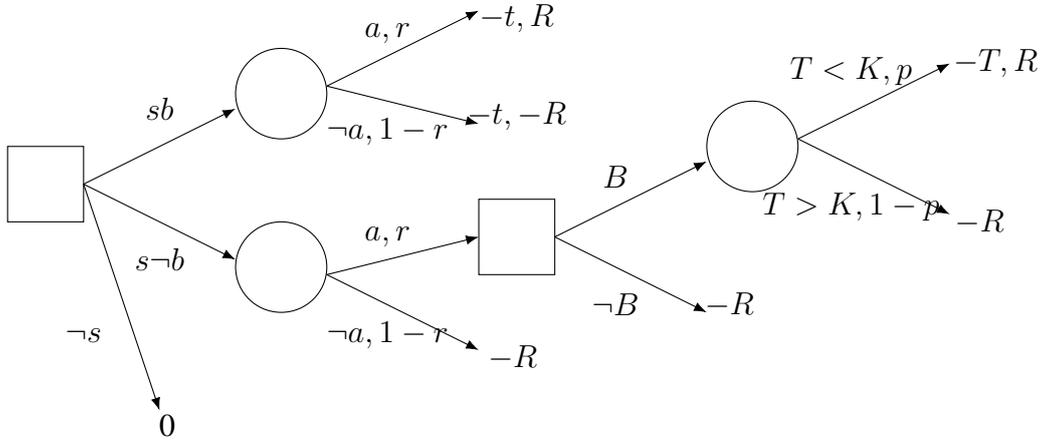
\begin{figure}[h]
   \begin{center}
    \begin{tikzpicture}

     \draw (0,9) rectangle (1,10);
     \draw (2,10.5) node {$sb$};
     \draw (2,8.5) node {$s\neg b$};
     \draw (1,7.5) node {$\neg s$};
     \draw[-{Latex}] (1,9.5) -- (3,10.5);
     \draw[-{Latex}] (1,9.5) -- (3,8.5);
     \draw[-{Latex}] (1,9.5) -- (2,6.5);
     \draw (2.1,6.3) node {0};
     \draw (3.6,10.7) circle[radius=6mm];
     \draw (5,11.5) node {$a,r$};
     \draw (5,10.2) node {$\neg a,1-r$};
     \draw (3.6,8.4) circle[radius=6mm];
     \draw (5,08.8) node {$a,r$};
     \draw (5,07.5) node {$\neg a,1-r$};
     \draw[-{Latex}] (4.2,10.8) -- (6.2,11.8);
     \draw[-{Latex}] (4.2,10.8) -- (6.2,10.3);
     \draw (6.7,11.7) node {$-t,R$};
     \draw (6.7,10.4) node {$-t,-R$};
     \draw (6.7,07.2) node {$-R~$};
     \draw[-{Latex}] (4.2,8.3) -- (6.2,7.3);
     \draw[-{Latex}] (4.2,8.3) -- (6.2,8.8);
     \draw (6.2,8.3) rectangle (7.2,9.3);
     \draw[-{Latex}] (7.2,8.8) -- (9.2,9.8);
     \draw[-{Latex}] (7.2,8.8) -- (9.2,7.8);
     \draw (8,09.6) node {$B$};
     \draw (8,07.9) node {$\neg B$};
     \draw (09.5,07.9) node {$-R$};
     \draw (9.8,10.0) circle[radius=6mm];
     \draw (11.1,11.0) node {$T<K,p$};
     \draw (11.1,09.2) node {$T>K,1-p$};
     \draw[-{Latex}] (10.4,10.1) -- (12.4,11.1);
     \draw[-{Latex}] (10.4,10.1) -- (12.4,9.1);
     \draw (13.0,11.1) node {$-T,R$};
     \draw (13.0,09.0) node {$-R~~~~$};

\end{tikzpicture}
\end{center}

   \caption{Alice's Decision Tree}\label{decisiontree01}
 \end{figure}

 There are three immediate possible actions ($\neg s$: not submit, $s\neg b$: submit and not buy the ticket, $sb$: submit and buy the ticket) and three possible scenarios ($a+$: paper accepted and sufficient budget, $a-$: paper accepted, but insufficient budget, $\neg a$: paper not accepted). Let's consider Alice's preferences: \\
  $\bullet$ $a+$: $sb\succ_{a+} s\neg b\succ_{a+}\neg s$ \\
  $\bullet$ $a-$: $sb\succ_{a-}\neg s\succ_{a-} s\neg b$ \\
  $\bullet$ $\neg a$: $\neg s\succ_{\neg a} s\neg b\succ_{\neg a} sb$.

 The interested reader can complete the exercice computing for which acceptance rate ($r$) and for which reward ($R$) Alice could take the risk to submit the paper and buy the ticket, but it is easy to observe that for low acceptance rates and for a risk averse decision maker the scenario $\neg a$ is by far the most likely to occur and that a rational decision will be to not submit the paper although this is extremely frustrating. \hfill $\Box$
\end{example}

\textbf{Discussion}. The previous example shows a typical problem of decision under risk. Once again the reader could ask why our framework improves the knowledge we have about these (well known) problems. There are essentially two reasons for which we consider our framework interesting.

1. The first is the more generally setting of the decision problem. In the example we do not use expected utilities, not even probabilities. We only show the potential actions, the different evaluation dimensions (in this case the scenarios), the outcomes and the preferences. Modelling the decision problem using the usual maximisation of subjective expected utility is certainly easy and useful, but only under well known hypotheses (which we do not discuss here). The decision problem however, is not to solve the expected utility maximisation, but to rank the potential actions considering the preferences along the different scenarios.

2. The second is related to the overall insatisfaction Alice will show with the solutions we prospect to her. Submitting the paper implies taking very high risks, while not submitting the paper (although rational) is very frustrating. This opens the question: can we design other options out of the ones we have presently? For this purpose we need to dig a little more in Alice's preferences and how these are computed. For the time being we only consider three dimensions corresponding to the three scenarios. On each of these the preferences are computed using a single dimension which is the balance between costs (the ticket price) and benefits (the reward). Potentially these are two distinct dimensions to consider (perhaps not to compensate as we did). Moreover, when considering the cost dimension we only consider the cost of buying the ticket and we do not distinguish (separate) the cost of booking the ticket without actually buying it. In other terms we do not consider the time between booking and buying the ticket, time which may have a value. If we do so we may realise that there is another potential action to do: pay (an amount $q$) for booking the ticket (fixing the price at $t$) and then wait to see if the paper is accepted or not. In other terms we can add at the root of the decision tree a new branch as shown in figure \ref{decisiontree02}. With that branch the preferences on the three scenarios turn to be: \\
  $\bullet$ $a+$: $sb\succ_{a+} sw\succ_{a+} s\neg b\succ_{a+}\neg s$ \\
  $\bullet$ $a-$: $sb\succ_{a-} sw\succ_{a-} \neg s\succ_{a-} s\neg b$ \\
  $\bullet$ $\neg a$: $\neg s\succ_{\neg a} sw\succ_{\neg a} s\neg b\succ_{\neg a} sb$. \\
The option $sw$ is always the second best one and a low cost $-q$ could compensate the high likelihood that the paper could not be accepted. With that model it is more likely that, although Alice is risk adverse, she will accept the risk wasting $q$ given the reward $R$ (which was not the case for $t$). What we observe is that representing the problem through our framework we get precise hints on how to construct new alternatives: separating new evaluation attributes. More precisely: separating the attribute ``cost'' in two different ones, cost of the booking and cost of the ticket, we are able to value the time between booking and buying and create new alternatives. In other terms we are able to introduce explicitly the time and the value of the information we get using this time.

\begin{figure}
\begin{center}
\begin{tikzpicture}

 \draw (0,9) rectangle (1,10);
 \draw (2.3, 9.7) node {$sw$};
 \draw (2.3, 8.5) node {$\neg s$};
 \draw (2.3,10.7) node {$sb$};
 \draw (2.3,11.7) node {$s\neg b$};
 \draw[-{Latex}] (1,9.5) -- (4,9.5);
 \draw[-{Latex}] (1,9.5) -- (2,8.5);
 \draw[-{Latex}] (1,9.5) -- (2,10.5);
 \draw[-{Latex}] (1,9.5) -- (2,11.5);
 \draw (4.6,9.5) circle[radius=6mm];
 \draw[-{Latex}] (5.2,9.5) -- (7,10.5);
 \draw[-{Latex}] (5.2,9.5) -- (7,8.5);
 \draw (6,10.3) node {$a$};
 \draw (6,08.4) node {$\neg a$};
 \draw (7.6,10.5) node {$-q,-t,R$};
 \draw (7.6,08.5) node {$-q~~~~$};
\end{tikzpicture}
\end{center}
  \caption{New option for Alice}\label{decisiontree02}
\end{figure}
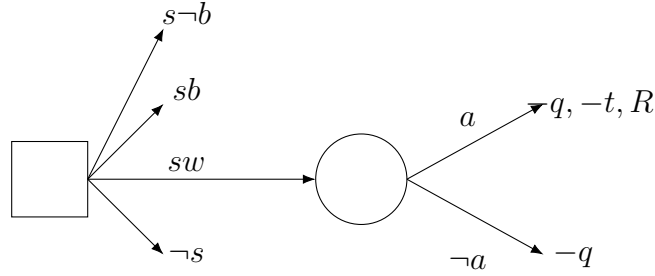

\section{Discussion}

Before concluding this paper it is important to discuss three points.

\begin{enumerate}
  \item Aiding to decide, constructing decision support, is a far more complex activity than solving a decision problem as presented here (see \cite{Rivett94}, \cite{rosen1book89}, \cite{MeinardTsoukiasEJOR2018}, \cite{MeinardTsoukias2022}, \cite{Tsoukias07aor}) and this is known since the very beginning of our discipline (see \cite{Ackoff1974}). The framework we suggest in this paper is not about conducting a decision aiding process, but about organising the formal knowledge we use when considering a client's demand in such a way that pushes us to focus upon the features of this demand before considering any solution method.
  \item The results we presented in this paper have some simple consequences: \\
       - The number of archetypal decision problems are finite and so are the possible archetypal methods that can be designed. We can construct many variations of any archetypal method, but our fundamental options in designing them are few and finite. \\
       - Real decision problems are sequences of different formal decision problems either because we refine the set of possible solutions or because we need to update beliefs, information and values or because we need to revise them. \\
       - Independently from modelling values, opinions or scenarios the formal basis of the process handling the problems is always the same: aggregating preferences expressed under different dimensions and then optimise for a given problem statement.
  \item It is easy to observe that while we try to provide decision support we alternate two types of activities: designing solutions and evaluating them (a concept already present in the literature; see \cite{simon57}, \cite{Hatchuel01}). In this paper we have shown that constructing a set of alternatives upon which apply a decision problem is itself a decision problem, but the creative process of constructing the set is a far more complex topic. We suspect that formal design theory could be useful for this purpose (see \cite{FerrettietalEJOR2018}, \cite{HatchuelWeil2009}, \cite{PluchinottaetalGDN2019}).
\end{enumerate}

\section{Conclusion}

The fundamental reason for which we considered reformulating what formally a decision problem is, is related to the training of decision analysts and operational researchers as well as the design of decision support methods. Most of the existing training frameworks are ``method-based'' in the sense that we describe ``methods'' without considering the specificities of our clients' problems. We suggest instead, that we should first characterise the problems and then find or construct an appropriate method. For this reason we suggest a new framework which does not make any reference on how problems are ``solved'', but on how problems are formally characterised with respect to the information provided by our clients.

We succeeded in establishing a minimal number of features (primitives) which are strictly necessary in order to elaborate a reasonable recommendation and at the same time are sufficient for characterising different archetypal decision problems. Such primitives are the set of alternatives, the problem statement, the descriptive and evaluative attributes and the preferences expressed by the clients. The different primitives being finite and their possible states being also finite means we have a finite number of archetypal decision problems and thus, a finite number of archetypal methods to solve them. We also show that constructing the set of alternatives (upon which we establish a decision problem) is a decision problem itself. The consequence is that a decision aiding process turns to be a sequence of decision problems to be solved where we alternate essentially two steps: aggregating preferences and optimising.

Our findings open several questions among which we want to emphasise two topics: \\
 1. The first one consists on how to organise training materials based on such a new framework and how to practically teach our discipline and our methods under this new perspective. \\
 2. The second one consists on establishing characterisation schemes between archetypal decision problems and archetypal decision support methods showing which properties are satisfied and why in order to provide explicable, accountable and convincing recommendations when automatic decision making devices are used.

We conclude with two open research questions: \\
 - if primitives are finite then it should be possible to establish protocols of minimal interaction with the client collecting the necessary information for handling a decision problem; \\
 - designing the set of alternatives is a crucial (and often neglected) step of the whole decision aiding process: for this purpose we need to establish new alternative design procedures and methods.

%

\end{document}